\documentclass[conference]{IEEEtran}
\pdfoutput=1

\IEEEoverridecommandlockouts
\usepackage{cite}
\usepackage{amsmath,amssymb,amsfonts}
\usepackage{algorithmic}
\usepackage{graphicx}
\usepackage{textcomp}
\usepackage{xcolor}

\usepackage{tabularx}
\usepackage{booktabs}
\usepackage{multirow}

\usepackage{caption}

\usepackage{CJKutf8}

\def\BibTeX{{\rm B\kern-.05em{\sc i\kern-.025em b}\kern-.08em
    T\kern-.1667em\lower.7ex\hbox{E}\kern-.125emX}}
\begin{document}

\title{Knowledge-injected Prompt Learning for Chinese Biomedical Entity Normalization
}

\author{
\IEEEauthorblockN{Songhua Yang}
\IEEEauthorblockA{\textit{School of Computer and Artificial Intelligence} \\
\textit{Zhengzhou University}\\
Zhengzhou, China \\
suprit@foxmail.com}
\and
\IEEEauthorblockN{Chenghao Zhang}
\IEEEauthorblockA{\textit{School of Computer and Artificial Intelligence} \\
\textit{Zhengzhou University}\\
Zhengzhou, China \\
zchcolorful@163.com.com}
\and
\IEEEauthorblockN{Hongfei Xu}
\IEEEauthorblockA{\textit{School of Computer and Artificial Intelligence} \\
\textit{Zhengzhou University}\\
Zhengzhou, China \\
hfxunlp@foxmail.com}
\and
\IEEEauthorblockN{Yuxiang Jia\textsuperscript{*}}
\IEEEauthorblockA{\textit{School of Computer and Artificial Intelligence} \\
\textit{Zhengzhou University}\\
Zhengzhou, China \\
ieyxjia@zzu.edu.cn}
\thanks{\textsuperscript{*}Corresponding author.}
}

\maketitle

\begin{abstract}
The Biomedical Entity Normalization (BEN) task aims to align raw, unstructured medical entities to standard entities, thus promoting data coherence and facilitating better downstream medical applications. Recently, prompt learning methods have shown promising results in this task. However, existing research falls short in tackling the more complex Chinese BEN task, especially in the few-shot scenario with limited medical data, and the vast potential of the external medical knowledge base has yet to be fully harnessed. To address these challenges, we propose a novel Knowledge-injected Prompt Learning (PL-Knowledge) method. Specifically, our approach consists of five stages: candidate entity matching, knowledge extraction, knowledge encoding, knowledge injection, and prediction output. By effectively encoding the knowledge items contained in medical entities and incorporating them into our tailor-made knowledge-injected templates, the additional knowledge enhances the model's ability to capture latent relationships between medical entities, thus achieving a better match with the standard entities. We extensively evaluate our model on a benchmark dataset in both few-shot and full-scale scenarios. Our method outperforms existing baselines, with an average accuracy boost of 12.96\% in few-shot and 0.94\% in full-data cases, showcasing its excellence in the BEN task.
\end{abstract}


\section{Introduction}

Biomedical Entity Normalization (BEN), also referred to as biomedical entity linking, is integral to aligning medical and biological terms with standardized entities. In the context of medicine, this process resolves ambiguity, vagueness, and misspellings, promoting uniformity in domain-specific textual terminology. With the exponential growth of medical data – encompassing electronic health records and scholarly medical literature – there lies a wealth of valuable knowledge for clinical decision-making, disease diagnosis, and patient management. However, the diversity and non-standardization of entities in medical texts, exacerbated by the intricacies of Chinese medical terminology and language, pose challenges for downstream tasks and applications. In response, BEN technology emerges to enhance medical entity standardization and facilitate efficient utilization of textual medical data.

\begin{figure}[t!]
\centering
\includegraphics[width=0.5\textwidth]{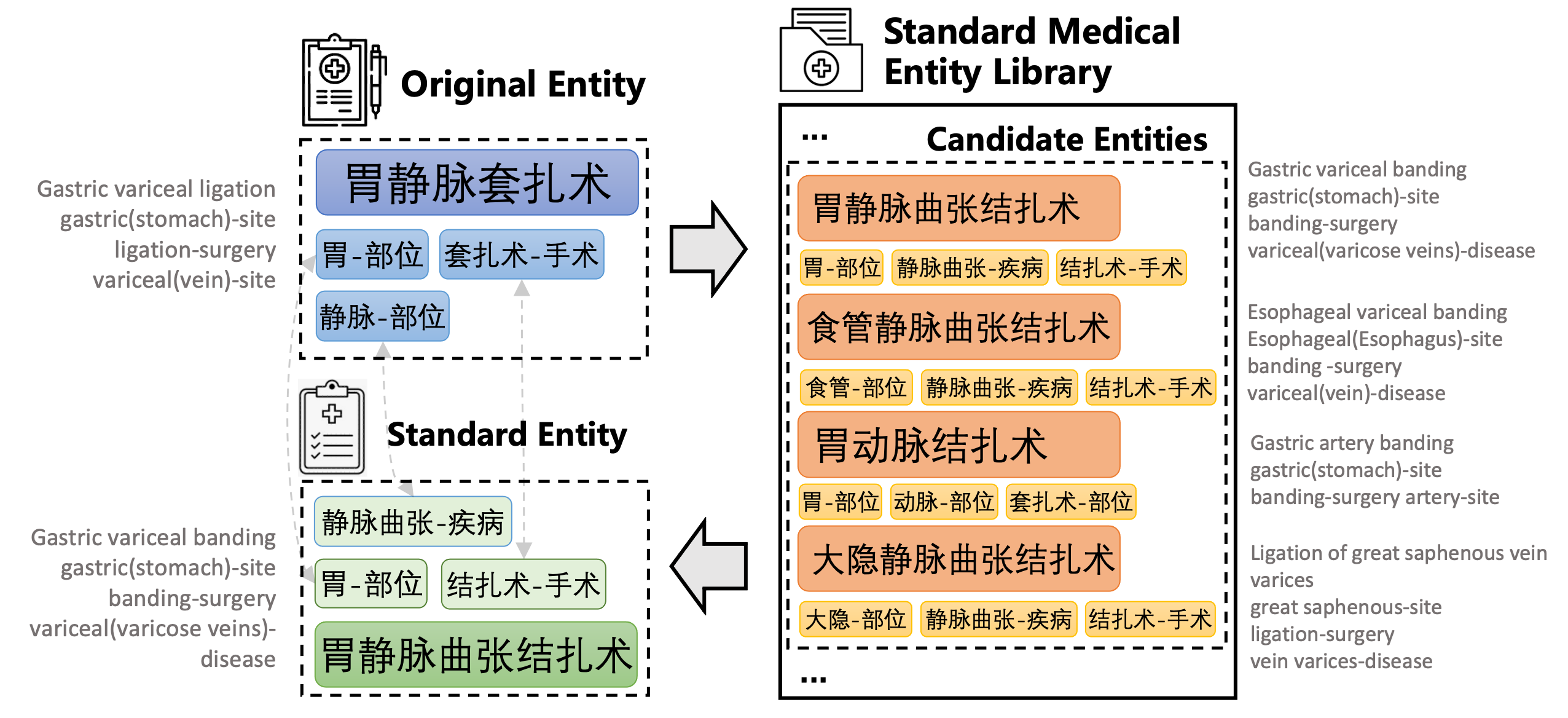}
\caption{An example of handling Chinese BEN task: the top left shows the non-standard original entity, which is matched through the standard library on the right. The correct standardized entity is output in the lower left. All medical knowledge  contained in the entity text listed alongside.}
\label{fig:stg1}
\end{figure}

While the field of BEN has garnered extensive attention over the years, its evolution has been shaped by advances in machine learning and natural language processing (NLP). Initially, rule-based approaches gauged textual similarity or curated dictionaries for entity matching, but struggled to scale. Machine learning methods split BEN into candidate word matching and classification stages, using classifiers and feature engineering. However, these methods faced limitations in semantic understanding and the need for extensive annotated data. Deep neural networks and word embedding algorithms subsequently improved BEN through enhanced generalization and semantic information capture. Pre-trained language models (PLMs), especially those trained on medical corpora, brought efficiency to BEN by refining task performance with minimal fine-tuning.

Despite notable strides, the complexity of biomedical entities, particularly in the Chinese context, persists as a challenge for BEN. The specialized nature of medical annotations incurs high costs, and the constant evolution of medical categories limits available annotated data. Thus, the importance of few-shot learning in BEN becomes evident, offering a pragmatic solution to real-world problems.

Prompt learning, a burgeoning paradigm in natural language processing, empowers PLMs to solve specific tasks using designed prompt templates. This approach enhances task understanding, reduces overfitting risks, and optimizes performance with scarce annotated data. While prompt learning has gained traction in biomedical entity normalization, Chinese BEN remains less explored, with untapped potential in leveraging external medical knowledge.

In this paper, we propose a novel Knowledge-injected Prompt Learning method (PL-Knowledge) to address challenges in Chinese BEN. Our approach involves candidate entity matching followed by prompt learning-based classification. The model comprises four stages: knowledge extraction, encoding, injection, and prediction. External medical knowledge refines entity embeddings, synergizing with PLMs through an intricate prompt template. Through extensive experiments on a benchmark Chinese BEN dataset, our PL-Knowledge method demonstrates substantial improvements, excelling in both few-shot and full-scale scenarios. Furthermore, our strategy underscores the crucial role of medical domain knowledge and our infusion strategy in enhancing model performance.

The main contributions of this work are as follows:

\indent \textbf{1)} We are the first to apply prompt learning to Chinese biomedical entity normalization task and demonstrate its effectiveness in few-shot scenarios.
\\
\indent \textbf{2)} We propose a knowledge-injected prompt learning method that leverages rich external medical domain knowledge base to enhance the performance of our model for BEN task.
\\
\indent \textbf{3)} Through experimental evaluation on a benchmark Chinese BEN dataset, our method achieves significant improvement over baseline methods in both few-shot and full-scale settings.

\section{Related Work}

The evolution of BEN techniques spans several decades, adapting to advancements in NLP and the burgeoning volume of biomedical text data. The approaches employed fall into distinct categories, including rule-based, machine learning-based, deep learning-based, and PLM-based methods. In earlier stages, rule-based methods were prevalent. Examples include MetaMap \cite{aronson2001effective}, which maintained dictionaries, and systems such as \cite{wermter2009high}, \cite{d2015sieve}, and \cite{kate2016normalizing}, which employed algorithms like edit distance and semantic similarity. While efficient, rule-based methods faced challenges in covering diverse cases and capturing nuanced information. Machine learning methods addressed these limitations by introducing candidate word matching and classification stages \cite{zhang2010entity}. Approaches like \cite{gaudan2005resolving} and \cite{leaman2013dnorm} harnessed SVMs and TF-IDF for entity resolution, leveraging Bag of Words and ranking learning. However, the insufficiency of labeled data and the incapability to grasp word semantics remained concerns. Deep neural network algorithms emerged next, leveraging word embedding techniques like Word2Vec and GloVe. Notable instances include \cite{li2017cnn}, \cite{wright2019normco}, and \cite{zhu2020latte}, deploying models such as CNNs and LSTMs to enhance entity resolution and coherence.

More recently, the advent of PLMs revolutionized BEN by enabling minimal fine-tuning. This encompassed approaches like \cite{xu2020generate}, retraining PLMs like BioBERT and ClinicalBERT \cite{alsentzer2019publicly, lee2020biobert}, and optimizing models for Chinese BEN \cite{yan2020knowledge, sui2022multi}. These methods exhibited substantial performance gains but had limitations in domain specificity and effective knowledge utilization. Prompt learning, a fourth paradigm in NLP, gained traction for few-shot and zero-shot tasks \cite{liu2023pre}. While some approaches integrated prompt learning with BEN \cite{zhang2021graphprompt, zhuenhancing, lai2022continuous}, they overlooked the complexities of Chinese BEN and the potential of external medical knowledge.

In summary, as BEN methods have evolved, limitations remain. Our proposed knowledge-injected prompt learning introduces innovative strategies to address biomedical entity normalization, bridging gaps in current methodologies.

\section{Approach}

In this section, we provide a detailed explanation of our proposed model. We begin by presenting a formal definition of the BEN task and then describe how we address this task in two stages, along with the specific processes and methods used in each stage.

\subsection{Problem Definition}

In the biomedical domain, an entity phrase $e$ may have multiple different expressions that are easily confused. To address this, there exists a standard entity name repository $E = \{e_1, e_2, ..., e_k\}$ containing the unique correct names for all entities. Given a list of $n$ biomedical entity phrases $O = \{o_1, o_2, ..., o_n\}$, the objective of the BEN task is to accurately match the correct standard term $s_j$ from the standard base $S$ for each original term $o_i$. Notably, numerous one-to-many, many-to-many, and many-to-one relationships exist between the original term $o$ and the standard term $s$, as shown in Table \ref{tab:sample}.

\begin{table*}[] 
\centering
\resizebox{0.99\textwidth}{!}{%
\begin{tabular}{l|l|l}   
\toprule
Original Entity & Standard Entity     & Relationship \\  
\midrule
\begin{tabular}[c]{@{}l@{}} 
\begin{CJK*}{UTF8}{gbsn}肝动脉化疗栓塞术\end{CJK*}\\  
 \textcolor{gray}{\scriptsize transcatheter arterial chemoembolization}\end{tabular} & 
\begin{tabular}[c]{@{}l@{}}
\begin{CJK*}{UTF8}{gbsn}经导管肝动脉栓塞术\#\#动脉化疗栓塞 \end{CJK*}\\
 \textcolor{gray}{\scriptsize transcatheter arterial embolization \#\# chemoembolization}
\end{tabular} & one to many  \\
\midrule
\begin{tabular}[c]{@{}l@{}}
\begin{CJK*}{UTF8}{gbsn}左侧腮腺肿块切除\#\#浅叶切除\#\#面神经松解减压术 \end{CJK*}\\ 
 \textcolor{gray}{\scriptsize left parotid mass excision \#\# superficial lobe excision} \\
 \textcolor{gray}{\scriptsize \#\# facial nerve decompression surgery}
\end{tabular} & 
\begin{tabular}[c]{@{}l@{}}
\begin{CJK*}{UTF8}{gbsn}面神经减压术\#\#腮腺病损切除术\end{CJK*} \\
 \textcolor{gray}{\scriptsize facial nerve decompression surgery \#\# parotid lesion excision}
\end{tabular} & many to many \\
\midrule
\begin{tabular}[c]{@{}l@{}}
\begin{CJK*}{UTF8}{gbsn}宫腔镜检查\#\#诊刮术 \end{CJK*}\\
 \textcolor{gray}{\scriptsize hysteroscopy \#\# diagnostic curettage}
\end{tabular} &
\begin{tabular}[c]{@{}l@{}}
\begin{CJK*}{UTF8}{gbsn}宫腔镜诊断性刮宫术\end{CJK*} \\
\textcolor{gray}{\scriptsize hysteroscopic diagnostic curettage}
\end{tabular} & many to one\\
\bottomrule
\end{tabular} 
}
\caption{Some representative examples of the relationship between the original entity and the standard entity.}
\label{tab:sample}
\end{table*}

\subsection{Preprocessing and Candidate Matching}

\begin{figure}[!h]
\centering
\includegraphics[width=0.5\textwidth]{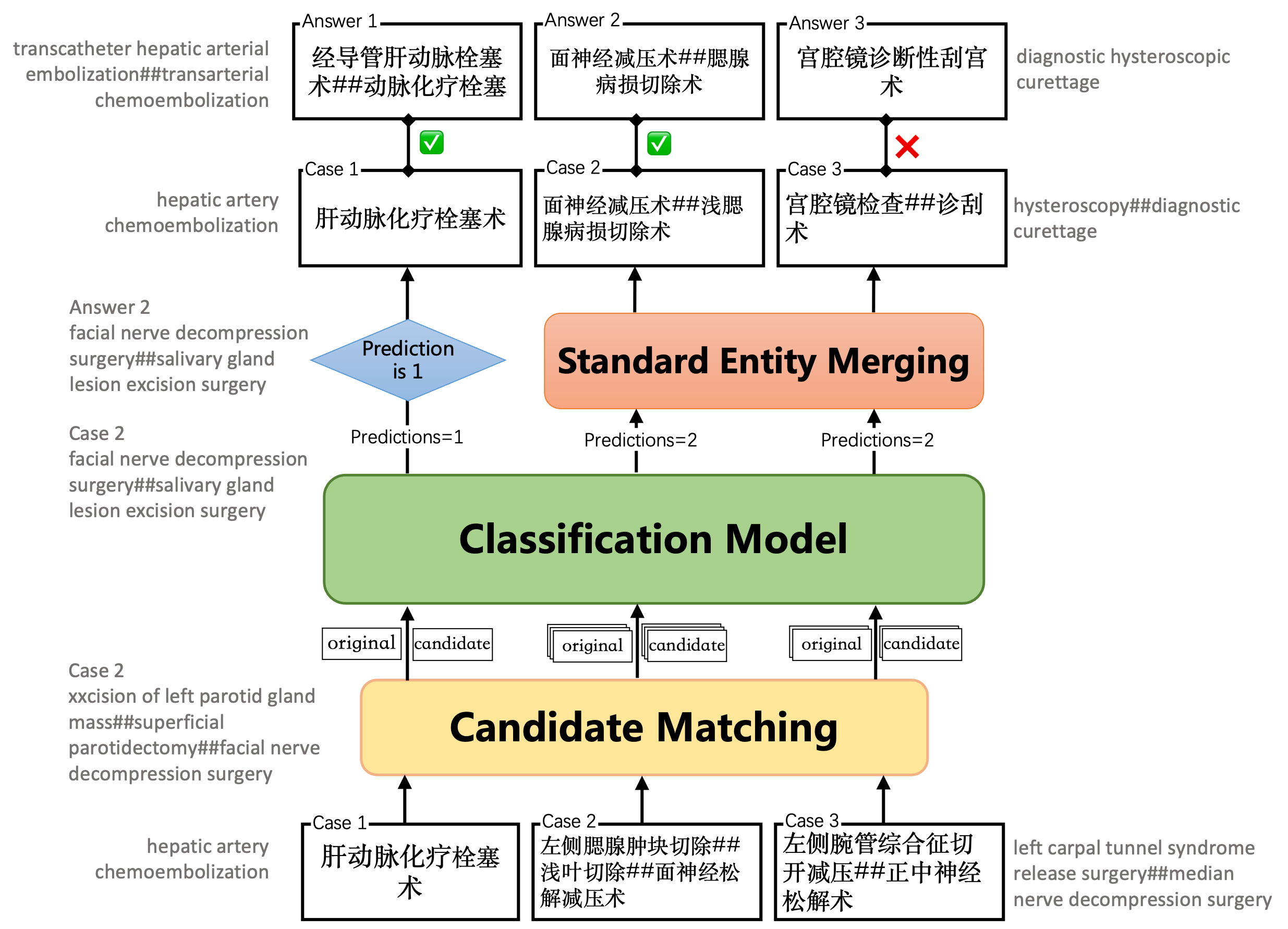}
\caption{Processing of representative examples during candidate term matching in the first stage. The three English text segments in the middle-left of the figure correspond to Case2.}
\label{fig:stg1}
\end{figure}

Medical professionals may use irregularities when writing original terms in real medical scenarios. Therefore, preprocessing of text data is necessary to improve the performance of entity normalization task. We manually remove irrelevant characters in the text, such as incorrect separators and punctuation marks. Moreover, in many-to-many situations between original and standard terms, delimiters between multiple entity words may be uncertain. We standardize them to easily split and merge entity phrases. We have devised a set of rules to flexibly handle many-to-many issues, as shown in Figure \ref{fig:stg1}. For the output of the classification model, if multiple standard terms are predicted, we merge the predicted result words and compare them with the standard terms.

Following the approach of previous studies \cite{li2017cnn, ji2020bert, zhuenhancing}, we divide the BEN task into two stages. In the first stage, we employ a text similarity algorithm to quickly match candidate entities. In particular, we segment all Chinese medical entities and calculate the Jaccard similarity coefficient between each original entity and each entity in the standard library. Based on the relevance scores, we select the top few entity phrases as candidate entities. The Jaccard similarity is a widely used text similarity calculation method that is simple, efficient, scalable, and insensitive to text length.

For each original entity $o$ and its corresponding candidate entities $C=\{c_1, c_2, ..., c_m\}$ in the dataset, we form a triplet $\{o, c_i, l\}$ as the input for the next classification model, where $l \in \{0, 1\}$ is the true label between $o$ and $c_i$. In one-to-many and many-to-many situations, the labels between the original entity containing multiple sub-words and the corresponding standard entity sub-words are set to 1. In this way, our model can ingeniously solve complex many-to-many situations through a binary classification approach.

\subsection{Prompt Learning}

\begin{figure*}[ht]
\centering
\includegraphics[width=1.0\textwidth]{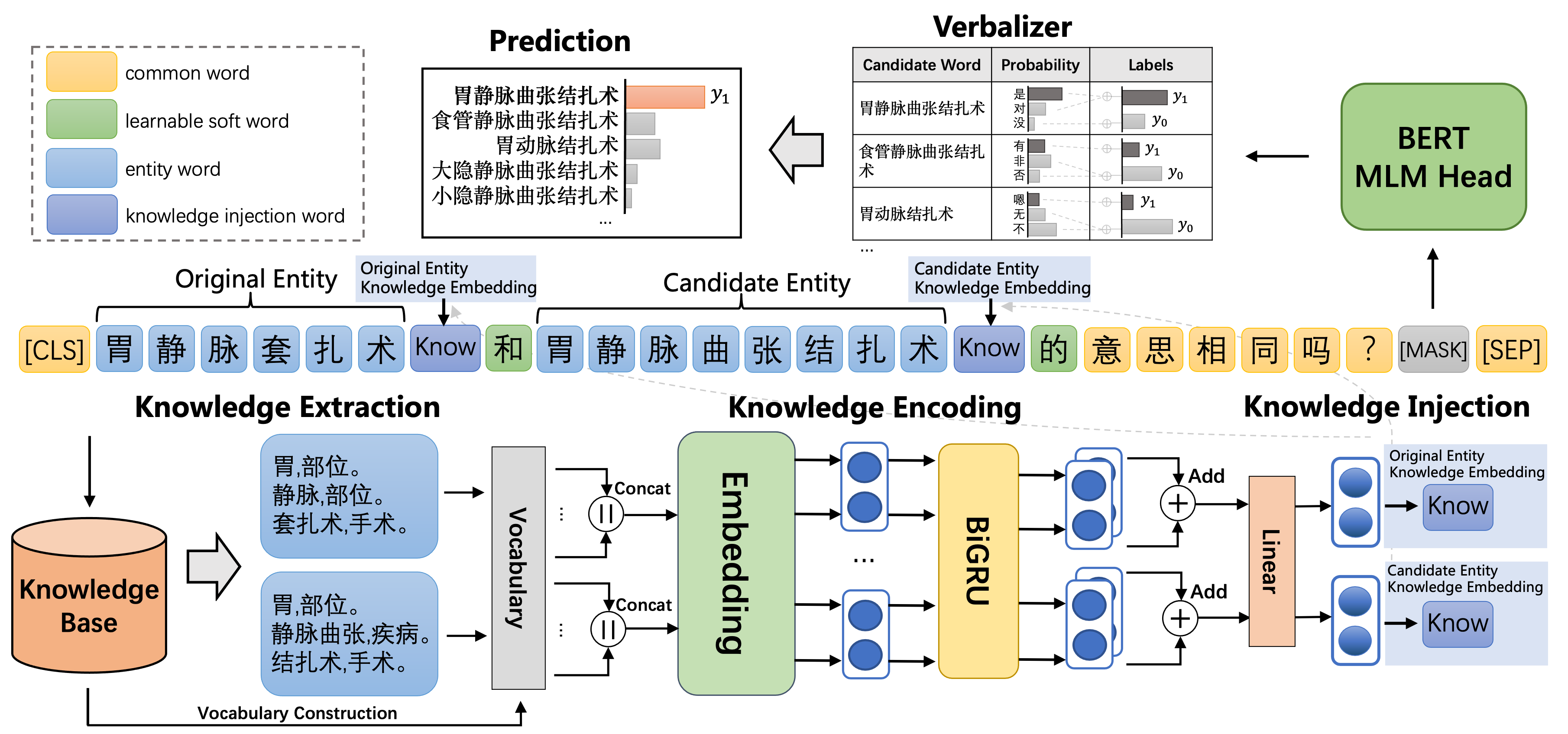}
\caption{The overall architecture of our model, featuring a carefully designed knowledge-injected prompt template in the middle. The knowledge extraction, knowledge encoding, and knowledge injection modules are shown moving from left to right at the bottom, while the final prediction output process is at the top.}
\label{fig:model}
\end{figure*}

We will now present a detailed explanation of our second-stage classification model based on prompt learning, as shown in Figure \ref{fig:model}.

\subsubsection{Overview}

The core idea of prompt learning is to construct a piece of natural language text (i.e., a template) and input it into the encoder of a Masked Language Model (MLM), converting the classification problem into a cloze-style task. Prompt learning models consist of a template and a verbalizer. Formally, given an original entity $o$ and its corresponding candidate entities $C=\{c_1,c_2,...,c_m\}$ as input, a tailor-made template $T(\cdot, \cdot)$ is used to create a text sequence $T(o,c)$, as shown in Table \ref{tab:templates}. The templates must include at least one [MASK] token. Let $\mathcal{M}$ be an MLM model, and $\mathcal{M}$ outputs the probability $P_{\mathcal{M}}\left([\mathrm{MASK}]=v \mid T(o,c)\right)\mid v \in \mathcal{V}$ of each word in the vocabulary $\mathcal{V}$ of $\mathcal{M}$ being filled in the [MASK] token.

The mapping from the predicted masked words to the labels indicating whether $o$ and $c$ are synonymous is called the verbalizer, which transforms the masked prediction task into a classification task. The performance of prompt learning is significantly influenced by the verbalizer's configuration. A verbalizer can be defined as $f: \mathcal{V}_y \mapsto \mathcal{Y}$, where $\mathcal{V}_y$ is the well-designed label words, a subset of the overall vocabulary $\mathcal{V}$, and $\mathcal{Y}$ represents the corresponding output labels (i.e., $y_0:0$ and $y_1:1$ for BEN). The final probability output of the classification model can be formulated as:

\begin{equation}
P(y \in \mathcal{Y} \mid T(o,c)) = g(P_{\mathcal{M}}([\mathrm{MASK}]=v \mid T(o,c) ) \mid v \in \mathcal{V}_y )
\end{equation}

where $g$ is the function corresponding to the verbalizer $f$, which calculates the final label probability score by the probability of label words in $\mathcal{V}_y$, thus obtaining the final classification result.

The prompt learning model's overall architecture and a representative example are illustrated in Figure \ref{fig:model}. The original term "\begin{CJK*}{UTF8}{gbsn}胃静脉套扎术\end{CJK*}"(gastric variceal ligation) and its corresponding candidate term ,"\begin{CJK*}{UTF8}{gbsn}胃静脉曲张结扎术\end{CJK*}"(gastric variceal banding) in this case, are inserted into the knowledge-injected template provided in Table \ref{tab:templates}. After converting the text into a vector representation and feeding it into the MLM $\mathcal{M}$, the probability distribution of label words for all candidate terms is obtained. The verbalizer decodes the output label probability distribution, where words like "\begin{CJK*}{UTF8}{gbsn}不\end{CJK*}", "\begin{CJK*}{UTF8}{gbsn}没\end{CJK*}" (no) correspond to $y_0$, while "\begin{CJK*}{UTF8}{gbsn}是\end{CJK*}", "\begin{CJK*}{UTF8}{gbsn}对\end{CJK*}" (yes) correspond to $y_1$. The candidate term with a label probability of $P(y_1)>P(y_0)$ is the final predicted standard term.

\subsubsection{Template Construction}

\begin{table*}[]
\small
\centering
\begin{tabular}{ll}
\toprule
\textbf{Template} & \textbf{Content} \\
\midrule
Manual Template & {[}Original Entity{]} \begin{CJK*}{UTF8}{gbsn}和\end{CJK*} {[}Candidate Entity{]} \begin{CJK*}{UTF8}{gbsn}意思相同吗？\end{CJK*} \\
& \textcolor{gray}{\small Is the meaning of {[}Original Entity{]} the same as {[}Candidate Entity{]}?} \\
\midrule
Mixed Template & {[}Original Entity{]} {[}soft{]} {[}soft:\begin{CJK*}{UTF8}{gbsn}和\end{CJK*}{]} {[}soft{]} {[}Candidate Entity{]} {[}soft: \begin{CJK*}{UTF8}{gbsn}的\end{CJK*}{]} \begin{CJK*}{UTF8}{gbsn}意思相同吗？\end{CJK*} \\
& \textcolor{gray}{\small Is the meaning of {[}Original Entity{]} {[}soft{]} {[}soft:and{]} {[}soft{]} {[}Candidate Entity{]} {[}soft: the{]} same?} \\
\midrule
Knowledgable Template & {[}Original Entity{]} {[}know{]} {[}soft:\begin{CJK*}{UTF8}{gbsn}和\end{CJK*}{]} {[}Candidate Entity{]} {[}know{]} {[}soft: \begin{CJK*}{UTF8}{gbsn}的\end{CJK*}{]} \begin{CJK*}{UTF8}{gbsn}意思相同吗？\end{CJK*} \\
& \textcolor{gray}{\small Is the meaning of {[}Original Entity{]} {[}know{]} {[}soft:and{]} {[}Candidate Entity{]} {[}know{]} {[}soft: the{]} same?} \\
\bottomrule
\end{tabular}
\caption{Demonstration of our three different types of prompt templates. "{[]}" denotes placeholders, where the original and candidate entities are filled in their respective positions. "{[}soft{]}" represents learnable soft tokens, with the subsequent characters indicating initialization with the corresponding vector; if not present, it is randomly initialized. "{[}know{]}" indicates that the vector at this position comes from knowledge injection.}
\label{tab:templates}
\end{table*}

The construction of templates is a critical factor in determining the performance of prompt learning. In order to maximize the ability of PLMs to perform classification task, we design a tailor-made knowledge-injected template, which can incorporates knowledge effectively. In addition, we also attempt the intuitive manual template and the automatically learnable mixed template for comparison. These templates are detailed in Table \ref{tab:templates}.

The manual template is the most straightforward type of template, containing only human-readable natural text. It adopts a hard-encoding approach, where the token embeddings in the template sequence are fixed to their corresponding vector representations. On the other hand, the mixed template combines both hard-encoding and soft-encoding approaches. Soft-encoding refers to encoding tokens as dynamic, learnable word vectors. For example, placeholders like [soft: \begin{CJK*}{UTF8}{gbsn}和\end{CJK*}] and [soft] in Table \ref{tab:templates} represent soft tokens. The embedding of the former token is initialized to the encoding vector corresponding to "\begin{CJK*}{UTF8}{gbsn}和\end{CJK*}" (and), while the latter token is randomly initialized. These soft token embeddings can be continuously optimized during the training process through gradient calculations. Soft tokens are more flexible and adaptive compared to hard-encoding, guiding the model to learn task-related key information and supporting more accurate classification. Several studies \cite{li2021prefix, liu2023pre, han2022ptr} have demonstrated that a prompt template with special learnable tokens can make prompt learning more effective. Therefore, we incorporate some soft-tokens in both Mixed Template and Knowledgeable Template to enhance model performance.

\subsubsection{External Knowledge Injection}

In this section, we will elaborate on the detailed process of prompt learning with knowledge injection, as shown in Figure \ref{fig:model}. We find that although medical entities may have some degree of irregular writing, they contain a wealth of medical knowledge, such as diseases, surgeries, and anatomical locations. This medical knowledge can act as crucial information to assist in matching original entities and standard entities. For example, in the case shown in the figure, we can see that both items contain "\begin{CJK*}{UTF8}{gbsn}胃, 部位\end{CJK*}"(stomach, site), "\begin{CJK*}{UTF8}{gbsn}静脉\end{CJK*}"(vein), and "\begin{CJK*}{UTF8}{gbsn}静脉曲张\end{CJK*}"(varicose veins) are related, and "\begin{CJK*}{UTF8}{gbsn}套扎术\end{CJK*}"(ligation procedure) and "\begin{CJK*}{UTF8}{gbsn}结扎术\end{CJK*}"(ligation procedure) are often synonymous. Therefore, we exploit a knowledge injection module to integrate medical knowledge from external knowledge base into the prompt template, which includes knowledge extraction, knowledge encoding, and knowledge injection.

We leverage the CMeKG \cite{ao2019Pre} Chinese medical knowledge graph, which contains knowledge about diseases, locations, and surgeries, to build an external knowledge base comprising 7,249 items denoted as $K=\{(k_1,r_1),(k_2,r_2),...,(k_n,r_n)\}$. To extract the corresponding knowledge item sequences $o \rightarrow o_k=\{(k_1,r_1),(k_2,r_2),...,(k_p,r_p)\}$ and $c \rightarrow c_k=\{(k_1,r_1),(k_2,r_2),...,(k_q,r_q)\}$ from the knowledge base $K$, we use the longest overlap matching method for the input original entity $o$ and the corresponding candidate entity $c$. To encode all knowledge items, we construct an additional vocabulary, treating each knowledge item and relationship type as an entity that can be encoded.

After encoding the vocabulary, we concatenate the two knowledge item sequences $o_k$ and $c_k$ to form a text sequence, which is then input into the subsequent embedding layer. For this layer, we use the encoder of a pre-trained medical domain PLM, MC-BERT \cite{zhang2020conceptualized}, as shown below:

\begin{flalign}
\centering
&h_o=Embedding((k_1,r_1)\parallel(k_2,r_2)\parallel...\parallel(k_p,r_p))
\\
&h_c=Embedding((k_1,r_1)\parallel(k_2,r_2)\parallel...\parallel(k_q,r_q))
\end{flalign}

To better capture the internal relationships and contextual information between knowledge item sequences, we incorporate a BiGRU module to refine the encoding of hidden vectors output by the Embedding layer. The BiGRU captures both forward and backward contextual information. We add the output forward and backward hidden state vectors $h'_{of}$, $h'_{ob}$, $h'_{cf}$, and $h'_{cb}$, and align the dimension of the resulting vector representation with the template through a Multi-Layer Perceptron (MLP). The final output vectors $h'_{ok}$ and $h'_{ck}$ are the knowledge encoding results to be injected into the template. The entire process can be formalized as follows:

\begin{flalign}
\centering
&h'_{of}, h'_{ob} = BiGRU(h_o), h'_{cf}, h'_{cb} = BiGRU(h_c)
\\
&h'_{ok} = MLP(h'_{of} + h'_{ob}), h'_{ck} = MLP(h'_{cf} + h'_{cb})
\end{flalign}

Finally, we substitute the knowledge embedding vectors $h'_{ok}$ and $h'_{ck}$ into the corresponding positions of the [know] placeholders in the template $T(o,c)$ near the original entity and candidate entity representations. In this way, the additional knowledge serves as a supplement to the entity information, assisting the model in making classification judgments.

\section{Experiments}

\subsection{Dataset}

To evaluate the effectiveness of our model, we conduct extensive experiments on the dataset provided by the Clinical Terminology Standardization Evaluation Task (CHIP 2019) \cite{huang2021CHIP2019}. . This publicly available BEN dataset is derived from actual Chinese electronic medical records and manually annotated using the "ICD9-2017 Clinical Edition" surgical standard terminology. The statistics are shown in Table \ref{tab:dataset}:

\begin{table}
\centering
\begin{tabular}{cccc}
\toprule
& Train & Dev & Test  \\
\midrule
one to one & 3596  & 897  & 1857  \\
one to many & 37  & 7 & 23  \\
many to many & 162   & 43 & 76 \\
many to one & 205  & 53 & 44\\
total & 4000  & 1000 & 2000  \\
\bottomrule
\end{tabular}
\caption{The statistics of CHIP2019 dataset. }
\label{tab:dataset}
\end{table}

To construct our external medical knowledge base, we utilize the Chinese Medical Knowledge Graph CMeKG \cite{ao2019Pre}, which is a large-scale knowledge graph containing 1.56 million conceptual relations and attribute triplets in the medical domain. We focus on extracting disease, location, and surgery-related knowledge items that may be relevant to the task dataset, and after filtering, screening, and deduplication, we obtain a total of 7,249 unique items. Given the complex relationships between the triples in the knowledge graph, we only retain the subject and predicate of each triple in our knowledge base.

\definecolor{mygreen}{RGB}{22,112,35}

\begin{table*}[ht]
\centering
\begin{tabularx}{\textwidth}{l*{8}{>{\raggedright\arraybackslash}X}}
\toprule
\multirow{2}{*}{\textbf{Model}} & 16-shot & 64-shot & 256-shot & 1024-shot & All(3500) \\
& Acc & Acc & Acc & Acc & Acc \\ \cmidrule(lr){1-6}
SVM(TF-IDF) & 23.87 & 29.56 & 37.35 & 42.21 & 47.45 \\
Edit-Distance & 24.87 & 30.47 & 38.05 & 44.18 & 48.36 \\
TextCNN & 25.34 & 37.57 & 52.21 & 76.46 & 86.12 \\ \cmidrule(lr){1-6}
BERT-based & 29.86 & 41.98 & 61.38 & 83.20 & 85.53\\
BERT-Knowledge & 25.87 & 45.52 & 62.16 & 84.10 & 87.04  \\ 
\cite{yan2020knowledge} & — & — & — & — & 89.30  \\ 
\cite{sui2022multi} & 32.47 & 48.29 & 68.79 & 88.44 & 91.14  \\ 
\cmidrule(lr){1-6}
PL-Manual & \textbf{59.83}{\color{mygreen}\scriptsize(+27.36)} & 64.24 & 74.79 & 88.35 & 90.67 \\
PL-Mixed & 54.39 & 61.94 & 73.59 & 88.20 & 90.83 \\
PL-Knowledge & 46.40 & \textbf{71.19}{\color{mygreen}\scriptsize(+22.90)} & \textbf{80.21}{\color{mygreen}\scriptsize(+11.42)} & \textbf{89.61}{\color{mygreen}\scriptsize(+2.17)} & \textbf{92.08}{\color{mygreen}\scriptsize(+0.94)} \\ 
\bottomrule
\end{tabularx}
\caption{Final experimental results of our model and baseline models on the CHIP 2019 BEN dataset. Bold in each column indicates the best result for that sample size. 16, 64, 256, and 1024 are the sample sizes; "All" refers to the full-scale evaluation, with accuracy as the evaluation metric for the models.}
\label{tab:results}
\end{table*}

\subsection{Baselines}

To provide a comprehensive evaluation of our model, we compare it to some baseline methods that cover a range of representative classification methods from machine learning, deep learning, to PLM-based models. Here is a brief description of each method:

SVM (TF-IDF): SVM \cite{cortes1995support} is a classic machine learning classification algorithm that is combined with the TF-IDF \cite{salton1983introduction} feature extraction method to vectorize text representation. This combination enables SVM to quickly capture key information in medical entities and provide more distinctive feature representations for classification task.

Edit-distance: A classic text similarity calculation method that can quickly calculate the similarity between two medical entities.

TextCNN: A Convolutional Neural Network (CNN) \cite{chen2015convolutional} suitable for text processing is used as a deep learning classification method baseline. TextCNN captures local contextual features between entity texts and ensures strong translation invariance.

BERT-based: A sequence classification method based on the pre-trained BERT model \cite{kenton2019bert}. Concatenating the original entity and standard entity into a sequence input model in a specified format for ranking.

BERT-Knowledge: In classification stage, concatenating the knowledge embedding with the two entities vectors with the last layer of BERT, and feeding this representation into a linear classifier. This is a common fine-tuning knowledge injection method \cite{elsherief2021latent} and is compared with our knowledge-injected prompt learning model.

\cite{yan2020knowledge}: Adopting an end-to-end sequence generative framework with category-based constraint decoding,  model-refining mechanism, and beam search techniques, achieving promising results in BEN task.

\cite{sui2022multi}: Using a multitask learning framework to simultaneously perform many-to-many prediction and entity classification subtasks, it is the previous best model to solve Chinese BEN task.

\subsection{Experimental Settings}

We employe the retrained MC-BERT\footnote{https://huggingface.co/freedomking/mc-bert} as the PLM for fine-tuning, prompt learning, and knowledge injection. Utilizing the Adam optimizer and a 1e-5 learning rate scheduler, we set 10 epochs, a batch size of 32, a maximum sequence length of 128, and a 0.3 dropout probability. We assess the model on the validation set after each epoch, selecting the best one for final testing on the test set. To evaluate performance in data-scarce scenarios, we conduct few-shot experiments with 16, 64, 256, and 1024 training samples, simulating real-world few-shot issues. We incorporate early stopping, terminating training if no significant improvement occurred within two epochs. For reliability, each test result was independently run five times using different random seeds, with the average value as the final result.

\begin{table}[t!]
\centering
\footnotesize
\begin{tabular}{l|lllll}
\toprule
\textbf{Model} & \textbf{16-shot} & \textbf{64-shot} & \textbf{256-shot} & \textbf{1024-shot} & \textbf{All(3500)} \\
\midrule
PL-K & 46.40 & 71.19 & 80.21 & 89.61 & 92.08 \\
\midrule
PL-K(w) & 44.35\textcolor{red}{\tiny{(-2.05)}} & 65.08\textcolor{red}{\tiny{(-6.11)}} & 76.47\textcolor{red}{\tiny{(-3.74)}} & 86.54\textcolor{red}{\tiny{(-3.07)}} & 90.79\textcolor{red}{\tiny{(-1.29)}} \\
\midrule
PL-K(r) & 42.17\textcolor{red}{\tiny{(-4.23)}} & 63.85\textcolor{red}{\tiny{(-7.34)}} & 73.60\textcolor{red}{\tiny{(-6.61)}} & 83.32\textcolor{red}{\tiny{(-6.29)}} & 88.23\textcolor{red}{\tiny{(-3.85)}} \\
\bottomrule
\end{tabular}
\caption{Comparison of experimental results for PL-Knowledge using different knowledge strategies.}
\label{tab:random_knowledge}
\end{table}

\subsection{Main Results}

\begin{table*}[t!]
\centering
\begin{tabular}{l|l}
\toprule
\multicolumn{1}{c|}{} & \multicolumn{1}{c}{\textbf{Template}} \\
\midrule
\#1   & {[}Original Entity{]}{[}know{]}{[}soft:\begin{CJK*}{UTF8}{gbsn}和\end{CJK*}{]}{[}Candidate Entity{]}{[}know{]}{[}soft:\begin{CJK*}{UTF8}{gbsn}的\end{CJK*}{]}\begin{CJK*}{UTF8}{gbsn}意思相同吗？\end{CJK*} \\[-0.5ex]
 & \scriptsize \textcolor{gray}{Is the meaning {[}soft:of{]} {[}Original Entity{]}{[}know{]} {[}soft:and{]} {[}Candidate Entity{]} {[}know{]} the same?} \\ 
\midrule
\#2   & {[}Original Entity{]}{[}know{]}\begin{CJK*}{UTF8}{gbsn}和\end{CJK*}{[}Candidate Entity{]} {[}know{]}\begin{CJK*}{UTF8}{gbsn}相同吗？\end{CJK*} \\[-0.5ex]
 & \scriptsize \textcolor{gray}{Is {[}Original Entity{]} {[}know{]} the same as {[}Candidate Entity{]} {[}know{]}?} \\
\midrule
\#3 & {[}Original Entity{]}{[}know{]}{[}soft:\begin{CJK*}{UTF8}{gbsn}和\end{CJK*}{]}{[}Candidate Entity{]}{[}know{]}{[}soft:\begin{CJK*}{UTF8}{gbsn}是\end{CJK*}{]}{[}soft:\begin{CJK*}{UTF8}{gbsn}不\end{CJK*}{]}{[}soft:\begin{CJK*}{UTF8}{gbsn}是\end{CJK*}{]}{[}soft:\begin{CJK*}{UTF8}{gbsn}相\end{CJK*}{]}{[}soft:\begin{CJK*}{UTF8}{gbsn}同\end{CJK*}{]}\begin{CJK*}{UTF8}{gbsn}？\end{CJK*} \\[-0.5ex]
 & \scriptsize \textcolor{gray}{{[}soft:Is{]} {[}Original Entity{]} {[}know{]} {[}soft:and{]} {[}Candidate Entity{]} {[}know{]} {[}soft:the{]} {[}soft:same{]} {[}soft:or{]} {[}soft:not{]}?} \\ 
\midrule
\#4 & \begin{CJK*}{UTF8}{gbsn}在医学术语中，\end{CJK*}{[}Original Entity{]}{[}know{]}{[}soft:\begin{CJK*}{UTF8}{gbsn}和\end{CJK*}{]}{[}Candidate Entity{]}{[}know{]}{[}soft:\begin{CJK*}{UTF8}{gbsn}的\end{CJK*}{]} \begin{CJK*}{UTF8}{gbsn}意思相同吗？\end{CJK*} \\[-0.5ex]
 & \scriptsize \textcolor{gray}{In medical terminology, is the meaning {[}soft:of{]} {[}Original Entity{]} {[}know{]} {[}soft:and{]} {[}Candidate Entity{]} {[}know{]} the same?} \\
\midrule
\#5   & \begin{CJK*}{UTF8}{gbsn}从专业医学的角度看，\end{CJK*}{[}Original Entity{]}{[}know{]}\begin{CJK*}{UTF8}{gbsn}与\end{CJK*}{[}Candidate Entity{]} {[}know{]}\begin{CJK*}{UTF8}{gbsn}是否指代同一概念？\end{CJK*}\\[-0.5ex]
 & \scriptsize \textcolor{gray}{From a professional medical , Do {[}Original Entity{]} {[}know{]} {[}soft:and{]} {[}Candidate Entity{]} {[}know{]} refer to the same concept?} \\ 
\bottomrule
\end{tabular}
\caption{Five different template settings. }
\label{tab:dif_templates}
\end{table*}

\begin{table}[t!]
\begin{tabular}{llllll}
\toprule
\multicolumn{1}{c}{\multirow{2}{*}{\textbf{Template}}} & 16-shot & 64-shot & 256-shot & 1024-shot & All(3500) \\
\multicolumn{1}{c}{} & Acc   & Acc   & Acc   & Acc   & Acc   \\ 
\midrule
\#1  & 46.40 & 71.19 & \textbf{80.21} & \textbf{89.61} & \textbf{92.08} \\ 
\midrule
\#2  & 45.57  &  69.34  & 78.83  & 84.68  & 91.23  \\ 
\midrule
\#3  & 52.25 & 67.53 & 78.01 & 87.48  & 91.19  \\ 
\midrule
\#4  & 47.51 & 71.37 &  79.18  & 88.25 & 91.39 \\ 
\midrule
\#5  & \textbf{53.73} & \textbf{72.28} & 79.35  & 89.03  & 91.61  \\ 
\bottomrule
\end{tabular}
\caption{Experimental results for different templates, with numbers corresponding to the template settings in Table \ref{tab:dif_templates}. Bold values in each column indicate the best experimental results.}
\label{tab:dif_template_res}
\end{table}

In this section, we present the experimental results of our model and various baseline methods on the CHIP 2019 BEN dataset, with Accuracy (Acc) used as the evaluation metric. Our prompt learning models based on manual templates, mixed templates containing soft tokens, and knowledge-injected templates are denoted as PL-Mannual, PL-Mixed, and PL-Knowledge, respectively. Based on the results presented in Table \ref{tab:results}, the following observations can be made:

\textbf{Prompt learning models significantly outperform baseline methods in few-shot scenarios.} As shown in the table, our prompt learning models consistently outperform the baseline methods in all-scale scenarios, especially in few-shot settings. For instance, our best prompt learning method achieves a notable improvement of 27.36\%, 22.90\%, and 11.42\% in the 16, 64, and 256-shot cases, respectively, which greatly surpasses the baseline methods. This result showcases the remarkable generalization capabilities of our prompt fine-tuning model in the few-shot learning task.

\textbf{Insufficient training data can impair the effectiveness of knowledge infusion.} We also observe that the effectiveness of knowledge infusion could be undermined by inadequate training data. In the 16-shot case, both the PL-Manual and BERT-based models outperformed the knowledge-injected PL-Knowledge and BERT-Knowledge. This implies that while the knowledge infusion strategy may enhance the model's generalization abilities, it may not be sufficient to counterbalance the negative effects of data scarcity when the training data is extremely limited.

\textbf{Knowledge enhancement greatly improves model performance in few-shot scenarios.} Our results demonstrate that, with the exception of the extremely scarce 16-shot case, knowledge-injected methods achieve notable performance in both few-shot and full-sample evaluation, indicating that the integration of external knowledge significantly enhances the model's understanding of the BEN task. The most significant improvement is observed in the 64-shot and 256-shot cases, suggesting that, with a moderately small sample size, the model can effectively incorporate the infused knowledge items and substantially enhance its performance. For the larger 1024-shot case, the performance improvement of knowledge-enhanced models is less pronounced.

\textbf{Our model maintains a slight advantage in all-data evaluation.} The results of the all-scale evaluation indicate that PL-Knowledge performs on par with the best baseline method in terms of performance, with a slight advantage.

\textbf{Template strategies have their pros and cons.} We observed that in the 1024-shot and full-scale cases, PL-Mixed outperformed PL-Manual, while in other cases, PL-Manual surpassed PL-Mixed. This implies that mixed templates incorporating soft tokens can provide a more flexible learning approach for the model, allowing it to better capture the correlation information between entities. However, for too few samples, soft tokens may not fully exploit their advantages, and PL-Manual may be better suited for few-shot scenarios. Therefore, in practical applications, the choice of template strategy should be based on factors such as sample size and data complexity.

\textbf{Baseline methods still hold reference value.} Among the baseline methods, \cite{sui2022multi} achieved the highest accuracy, demonstrating their excellent performance on the BEN task. Although the accuracy of early TF-IDF, Edit-Distance, and TextCNN methods was relatively low, they provided an important benchmark for evaluating the performance differences between our methods and traditional machine learning and neural network methods. 

In summary, the experimental results demonstrate that our model achieves good performance in various scenarios, particularly in the few-shot learning task, proving the strong generalization capabilities and practicality of our method.

\subsection{Impact of Knowledge Injection Strategies}

We further investigated the impact of different knowledge injection strategies on model performance by modifying the embedding method of knowledge items in the PL-Knowledge model to use Word2vec and random initialization. Med-word2vec \cite{MedicineWord2vec} is an early word vector table pre-trained on a moderately sized Chinese medical corpus, with weaker representation capabilities in the medical domain compared to MC-BERT. On the other hand, random initialization does not introduce any actual knowledge, requiring the model to learn on its own. Therefore, we can observe the impact on model performance when introducing limited knowledge and not introducing it at all.

As shown in Table \ref{tab:random_knowledge}, when the knowledge embedding method adopts Word2vec and random initialization, the model performance declines in both cases, especially with random initialization. This phenomenon confirms the effectiveness of our knowledge injection strategy in the entity normalization task, as external knowledge can help the model capture associations between entities, thereby improving the model's generalization ability in this task.

Furthermore, this result also reveals the potential negative impact of randomly initialized knowledge embeddings on model performance. Randomly initialized embeddings cannot provide effective knowledge information to the model, causing a certain degree of disturbance during the learning process. Compared to MC-BERT, word2vec has a weaker representation capability.

\subsection{Analysis of Different Templates}

In this section, we analyze and compare the impact of different template settings in the PL-Knowledge method on the performance of the BEN task. A well-designed template can help PLMs better understand task requirements and improve model performance. In addition to the three token types of hard, soft, and knowledge-injected, the length and specific statements in the template are also essential factors in prompt learning. Appropriate template length can significantly improve model performance, and some specific statements can better induce PLMs to output correct results. These factors need to be experimented with and adjusted based on specific tasks and datasets \cite{li2021prefix, liu2023pre, lester2021power, gao2021making}.

Considering the above factors, we designed five different templates in Table \ref{tab:dif_templates} based on token type, template length, and specific statements. Table \ref{tab:dif_template_res} shows the experimental results of different templates under various sample sizes. Although each template performs well in specific scenarios, Template \#1 consistently achieves the highest accuracy in the 256-shot, 1024-shot, and full-scale scenarios. On the other hand, Template \#5 performs best in the small sample scenarios of 16-shot and 64-shot, indicating that this template is better suited for such situations. Notably, although both \#4 and \#5 templates contain similar specific prefixes, their performance differences are significant, further confirming the significant impact of specific statements on model performance. In all scenarios, the completely manual \#2 template and the entirely soft token-based \#3 template perform relatively poorly, emphasizing the need for the appropriate incorporation of soft tokens to improve prompt fine-tuning performance. These findings provide essential guidance for selecting appropriate template strategies and directions in practical applications.

\section{Conclusion}

In this study, we addressed the challenges of the Chinese BEN task, focusing on data-limited few-shot scenarios. Our proposed PL-Knowledge method combines an external medical knowledge base with prompt learning to enhance BEN performance. Empirical results on a benchmark Chinese BEN dataset showcase the efficacy of our model, outperforming existing baselines in both few-shot and full-data scenarios. Through ablation and comparison experiments, we highlight the pivotal role of knowledge injection and template design in boosting model performance.

Future work will delve into more effective strategies for integrating external knowledge to further enhance task performance, and will explore maintaining model effectiveness in zero-shot scenarios. Additionally, our focus on the Chinese context limits our exploration. In forthcoming research, we aim to validate our model on English datasets and bridge this gap in our analysis.

\bibliography{references}

\begin{thebibliography}{10}
\providecommand{\url}[1]{#1}
\csname url@samestyle\endcsname
\providecommand{\newblock}{\relax}
\providecommand{\bibinfo}[2]{#2}
\providecommand{\BIBentrySTDinterwordspacing}{\spaceskip=0pt\relax}
\providecommand{\BIBentryALTinterwordstretchfactor}{4}
\providecommand{\BIBentryALTinterwordspacing}{\spaceskip=\fontdimen2\font plus
\BIBentryALTinterwordstretchfactor\fontdimen3\font minus
  \fontdimen4\font\relax}
\providecommand{\BIBforeignlanguage}[2]{{%
\expandafter\ifx\csname l@#1\endcsname\relax
\typeout{** WARNING: IEEEtran.bst: No hyphenation pattern has been}%
\typeout{** loaded for the language `#1'. Using the pattern for}%
\typeout{** the default language instead.}%
\else
\language=\csname l@#1\endcsname
\fi
#2}}
\providecommand{\BIBdecl}{\relax}
\BIBdecl

\bibitem{aronson2001effective}
A.~R. Aronson, ``Effective mapping of biomedical text to the umls
  metathesaurus: the metamap program.'' in \emph{Proceedings of the AMIA
  Symposium}.\hskip 1em plus 0.5em minus 0.4em\relax American Medical
  Informatics Association, 2001, p.~17.

\bibitem{wermter2009high}
J.~Wermter, K.~Tomanek, and U.~Hahn, ``High-performance gene name normalization
  with geno,'' \emph{Bioinformatics}, vol.~25, no.~6, pp. 815--821, 2009.

\bibitem{d2015sieve}
J.~D’Souza and V.~Ng, ``Sieve-based entity linking for the biomedical
  domain,'' in \emph{Proceedings of the 53rd Annual Meeting of the Association
  for Computational Linguistics and the 7th International Joint Conference on
  Natural Language Processing (Volume 2: Short Papers)}, 2015, pp. 297--302.

\bibitem{kate2016normalizing}
R.~J. Kate, ``Normalizing clinical terms using learned edit distance
  patterns,'' \emph{Journal of the American Medical Informatics Association},
  vol.~23, no.~2, pp. 380--386, 2016.

\bibitem{zhang2010entity}
W.~Zhang, J.~Su, C.~L. Tan, and W.~T. Wang, ``Entity linking leveraging
  automatically generated annotation,'' in \emph{Proceedings of the 23rd
  International Conference on Computational Linguistics (Coling 2010)}, 2010,
  pp. 1290--1298.

\bibitem{gaudan2005resolving}
S.~Gaudan, H.~Kirsch, and D.~Rebholz-Schuhmann, ``Resolving abbreviations to
  their senses in medline,'' \emph{Bioinformatics}, vol.~21, no.~18, pp.
  3658--3664, 2005.

\bibitem{leaman2013dnorm}
R.~Leaman, R.~Islamaj~Do{\u{g}}an, and Z.~Lu, ``Dnorm: disease name
  normalization with pairwise learning to rank,'' \emph{Bioinformatics},
  vol.~29, no.~22, pp. 2909--2917, 2013.

\bibitem{li2017cnn}
H.~Li, Q.~Chen, B.~Tang, X.~Wang, H.~Xu, B.~Wang, and D.~Huang, ``Cnn-based
  ranking for biomedical entity normalization,'' \emph{BMC bioinformatics},
  vol.~18, pp. 79--86, 2017.

\bibitem{wright2019normco}
D.~Wright, \emph{NormCo: Deep disease normalization for biomedical knowledge
  base construction}.\hskip 1em plus 0.5em minus 0.4em\relax University of
  California, San Diego, 2019.

\bibitem{zhu2020latte}
M.~Zhu, B.~Celikkaya, P.~Bhatia, and C.~K. Reddy, ``Latte: Latent type modeling
  for biomedical entity linking,'' in \emph{Proceedings of the AAAI Conference
  on Artificial Intelligence}, vol.~34, no.~05, 2020, pp. 9757--9764.

\bibitem{xu2020generate}
D.~Xu, Z.~Zhang, and S.~Bethard, ``A generate-and-rank framework with semantic
  type regularization for biomedical concept normalization,'' in
  \emph{Proceedings of the 58th Annual Meeting of the Association for
  Computational Linguistics}, 2020, pp. 8452--8464.

\bibitem{alsentzer2019publicly}
E.~Alsentzer, J.~R. Murphy, W.~Boag, W.-H. Weng, D.~Jin, T.~Naumann,
  W.~Redmond, and M.~B. McDermott, ``Publicly available clinical bert
  embeddings,'' \emph{NAACL HLT 2019}, p.~72, 2019.

\bibitem{lee2020biobert}
J.~Lee, W.~Yoon, S.~Kim, D.~Kim, S.~Kim, C.~H. So, and J.~Kang, ``Biobert: a
  pre-trained biomedical language representation model for biomedical text
  mining,'' \emph{Bioinformatics}, vol.~36, no.~4, pp. 1234--1240, 2020.

\bibitem{yan2020knowledge}
J.~Yan, Y.~Wang, L.~Xiang, Y.~Zhou, and C.~Zong, ``A knowledge-driven
  generative model for multi-implication chinese medical procedure entity
  normalization,'' in \emph{Proceedings of the 2020 Conference on Empirical
  Methods in Natural Language Processing (EMNLP)}, 2020, pp. 1490--1499.

\bibitem{sui2022multi}
X.~Sui, K.~Song, B.~Zhou, Y.~Zhang, and X.~Yuan, ``A multi-task learning
  framework for chinese medical procedure entity normalization,'' in
  \emph{ICASSP 2022-2022 IEEE International Conference on Acoustics, Speech and
  Signal Processing (ICASSP)}.\hskip 1em plus 0.5em minus 0.4em\relax IEEE,
  2022, pp. 8337--8341.

\bibitem{liu2023pre}
P.~Liu, W.~Yuan, J.~Fu, Z.~Jiang, H.~Hayashi, and G.~Neubig, ``Pre-train,
  prompt, and predict: A systematic survey of prompting methods in natural
  language processing,'' \emph{ACM Computing Surveys}, vol.~55, no.~9, pp.
  1--35, 2023.

\bibitem{zhang2021graphprompt}
J.~Zhang, Z.~Wang, S.~Zhang, M.~M. Bhalerao, Y.~Liu, D.~Zhu, and S.~Wang,
  ``Graphprompt: Biomedical entity normalization using graph-based prompt
  templates,'' \emph{arXiv preprint arXiv:2112.03002 , year={2021}}.

\bibitem{zhuenhancing}
T.~Zhu, Y.~Qin, Q.~Chen, B.~Hu, and Y.~Xiang, ``Enhancing entity
  representations with prompt learning for biomedical entity linking,'' 2022.

\bibitem{lai2022continuous}
Z.~Lai, B.~Fu, S.~Wei, and X.~Shi, ``Continuous prompt enhanced biomedical
  entity normalization,'' in \emph{Natural Language Processing and Chinese
  Computing: 11th CCF International Conference, NLPCC 2022, Guilin, China,
  September 24--25, 2022, Proceedings, Part II}.\hskip 1em plus 0.5em minus
  0.4em\relax Springer, 2022, pp. 61--72.

\bibitem{ji2020bert}
Z.~Ji, Q.~Wei, and H.~Xu, ``Bert-based ranking for biomedical entity
  normalization,'' \emph{AMIA Summits on Translational Science Proceedings},
  vol. 2020, p. 269, 2020.

\bibitem{li2021prefix}
X.~L. Li and P.~Liang, ``Prefix-tuning: Optimizing continuous prompts for
  generation,'' in \emph{Proceedings of the 59th Annual Meeting of the
  Association for Computational Linguistics and the 11th International Joint
  Conference on Natural Language Processing (Volume 1: Long Papers)}, 2021, pp.
  4582--4597.

\bibitem{han2022ptr}
X.~Han, W.~Zhao, N.~Ding, Z.~Liu, and M.~Sun, ``Ptr: Prompt tuning with rules
  for text classification,'' \emph{AI Open}, vol.~3, pp. 182--192, 2022.

\bibitem{ao2019Pre}
Y.~Ao and Zan, ``Preliminary study on the construction of chinese medical
  knowledge graph,'' \emph{JOURNAL OF CHINESE INFORMATION PROCESSING}, vol.~33,
  no.~10, p.~9, 2019.

\bibitem{zhang2020conceptualized}
N.~Zhang, Q.~Jia, K.~Yin, L.~Dong, F.~Gao, and N.~Hua, ``Conceptualized
  representation learning for chinese biomedical text mining,'' \emph{arXiv
  preprint arXiv:2008.10813}, 2020.

\bibitem{huang2021CHIP2019}
Huang, Jiao, Tang, Chen, and Yanjun, ``Overview of the chip2019 shared task
  track1 : Normalizationof chinese clinical terminology,'' \emph{JOURNAL OF
  CHINESE INFORMATION PROCESSING}, vol.~35, no.~3, p.~6, 2021.

\bibitem{cortes1995support}
C.~Cortes and V.~Vapnik, ``Support-vector networks,'' \emph{Machine learning},
  vol.~20, no.~3, pp. 273--297, 1995.

\bibitem{salton1983introduction}
G.~Salton and M.~J. McGill, \emph{Introduction to modern information
  retrieval}.\hskip 1em plus 0.5em minus 0.4em\relax McGraw-Hill, 1983.

\bibitem{chen2015convolutional}
Y.~Chen, ``Convolutional neural network for sentence classification,'' Master's
  thesis, University of Waterloo, 2015.

\bibitem{kenton2019bert}
J.~D. M.-W.~C. Kenton and L.~K. Toutanova, ``Bert: Pre-training of deep
  bidirectional transformers for language understanding,'' in \emph{Proceedings
  of NAACL-HLT}, 2019, pp. 4171--4186.

\bibitem{elsherief2021latent}
M.~ElSherief, C.~Ziems, D.~Muchlinski, V.~Anupindi, J.~Seybolt,
  M.~De~Choudhury, and D.~Yang, ``Latent hatred: A benchmark for understanding
  implicit hate speech,'' in \emph{Proceedings of the 2021 Conference on
  Empirical Methods in Natural Language Processing}, 2021, pp. 345--363.

\bibitem{MedicineWord2vec}
Y.~Weng, ``Chineseword2vecmedicine,''
  \url{https://github.com/WENGSYX/Chinese-Word2vec-Medicine}, 2021.

\bibitem{lester2021power}
B.~Lester, R.~Al-Rfou, and N.~Constant, ``The power of scale for
  parameter-efficient prompt tuning,'' in \emph{Proceedings of the 2021
  Conference on Empirical Methods in Natural Language Processing}, 2021, pp.
  3045--3059.

\bibitem{gao2021making}
T.~Gao, A.~Fisch, and D.~Chen, ``Making pre-trained language models better
  few-shot learners,'' in \emph{Joint Conference of the 59th Annual Meeting of
  the Association for Computational Linguistics and the 11th International
  Joint Conference on Natural Language Processing, ACL-IJCNLP 2021}.\hskip 1em
  plus 0.5em minus 0.4em\relax Association for Computational Linguistics (ACL),
  2021, pp. 3816--3830.

\end{thebibliography}
\bibliographystyle{IEEEtran}

\end{document}